\documentclass[conference]{IEEEtran}
\IEEEoverridecommandlockouts
\usepackage{cite}
\usepackage{amsmath,amssymb,amsfonts}

\usepackage{graphicx}
\usepackage{textcomp}
\usepackage{xcolor}
\usepackage{booktabs}
\usepackage{multirow}
\usepackage{amsmath,amssymb}

\usepackage{tikz}
\usepackage{algorithm}
\usepackage{algorithmic}
\usepackage{url}
\usepackage{color}
\usepackage{colortbl}
  \usetikzlibrary{positioning,fit,backgrounds,arrows.meta}
\def\BibTeX{{\rm B\kern-.05em{\sc i\kern-.025em b}\kern-.08em
    T\kern-.1667em\lower.7ex\hbox{E}\kern-.125emX}}
\begin{document}

\title{Mamba-based Selective State Space Modeling Improves the Accuracy-Complexity Tradeoff of SmolVLA Vision-Language-Action Experts 
\thanks{}
}



\author{\IEEEauthorblockN{Farida Mohsen$^1$, Thowayba Elkaffash$^{1,2}$, Mohammad Reza Chalak Qazani$^3$, \\
Mohamed Mabrok$^2$, Nader Meskin$^2$, Ali Safa$^1$}
\IEEEauthorblockA{$^1$\textit{College of Science and Engineering, Hamad Bin Khalifa University, Doha, Qatar} \\
$^2$\textit{College of Engineering, Qatar University, Doha, Qatar}\\
$^3$\textit{College of Science and Engineering, James Cook University, Townsville, QLD, 4814, Australia}\\
fmohsen@hbku.edu.qa
}}

\maketitle

\begin{abstract}
Vision-language-action (VLA) models face a crucial tradeoff between their task success rate and the policy-call frequency. Executing a single action per inference ($N=1$) enables  accurate robot control but comes at the cost of huge compute time overheads, making real-time implementation infeasible. On the other hand, executing longer action horizons before replanning ($N\gg1$) reduces compute complexity, but inevitably degrades the system's success rate. In order to improve the VLA accuracy-complexity tradeoff, this paper investigates Mamba's selective state-space modeling as an alternative to causal self-attention within the action expert of the popular SmolVLA model, widely used as a reference model for its highly accurate yet low complexity nature. 
We evaluate both the Mamba- and Transformer-based experts on the widely-adopted LIBERO benchmark suites across three execution horizons $N\!\in\!\{1,25,50\}$, respectively corresponding to high, moderate and low compute complexities. Our results remarkably show that the advantage of the Mamba expert increases with the execution horizon, indicating significant success retention under long execution horizons $N = 50$ and $N = 25$. When $N = 50$ actions are executed before replanning (i.e., corresponding to feasible real-time deployment), the Mamba expert outperforms the Transformer baseline by $7.8\%$. In addition, when $N = 25$ actions are executed before replanning, our Mamba expert outperforms the Transformer baseline by $3.7\%$. Finally, under per-action replanning ($N=1$), our Mamba variant matches the Transformer-based mean success rate while significantly reducing the overall model parameter complexity by $24\%$ thanks to Mamba's compute-efficient nature. 

\end{abstract}


\section{Introduction}
\begin{figure}[t]
\centering
\includegraphics[width=\columnwidth]{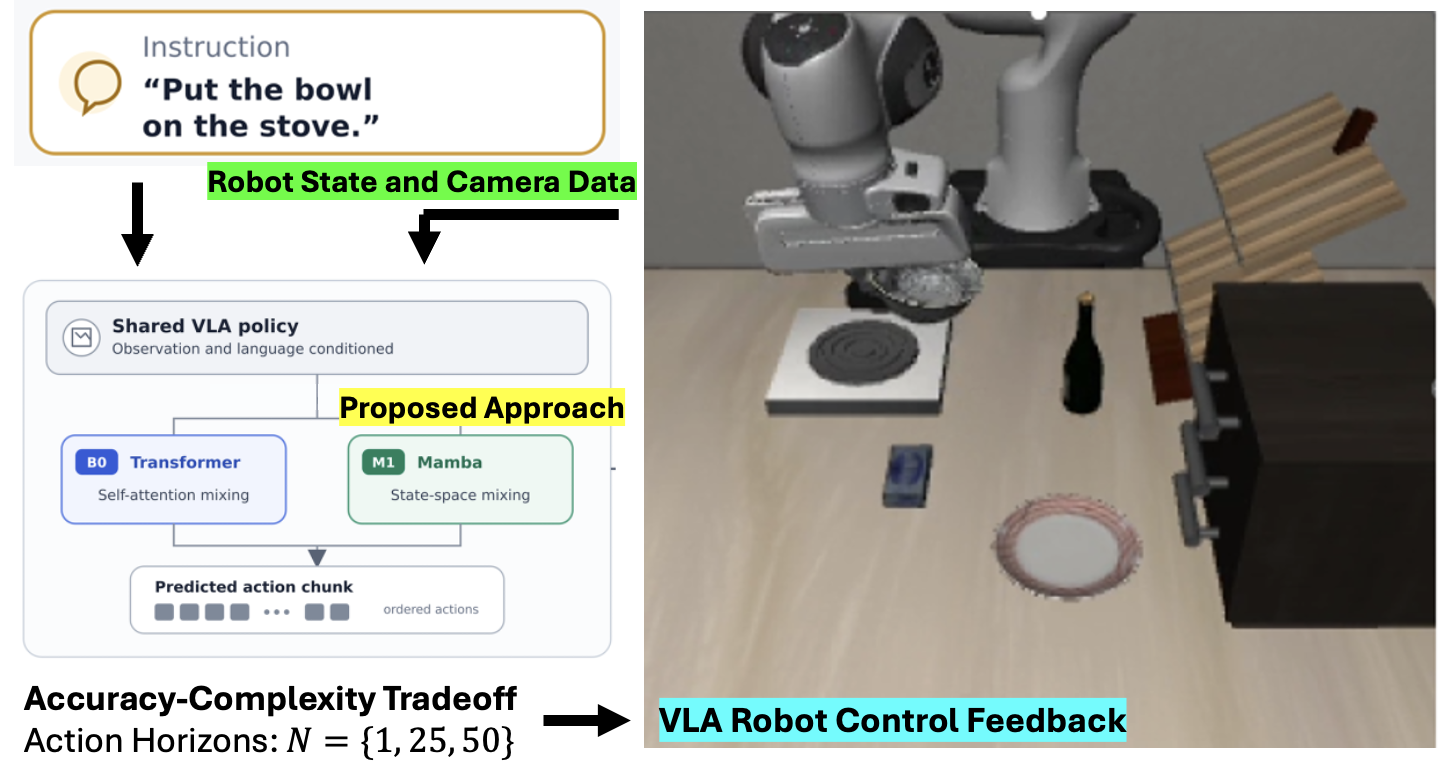}
\caption{This work provides a novel investigation of the use of Mamba-type Selective State Space Modeling for Vision-Language-Action (VLA) models. Text instructions, as well as robot state and camera data are fed to two VLA setups: a Transformer-based reference and our proposed Mamba-based model. The generated output actions are then used to control a robot arm in a standard benchmarking simulation environment. Remarkably, this work demonstrates that using Mamba leads to significant gains (up to $+7.8\%$) in success rate while using longer action horizons $N$, thereby enabling a better accuracy-complexity tradeoff compared to Transformers (which is crucial for feasible real-time system deployment).}
\label{fig:retention}
\end{figure}

\begin{figure*}[]
\centering
\includegraphics[width=\textwidth]{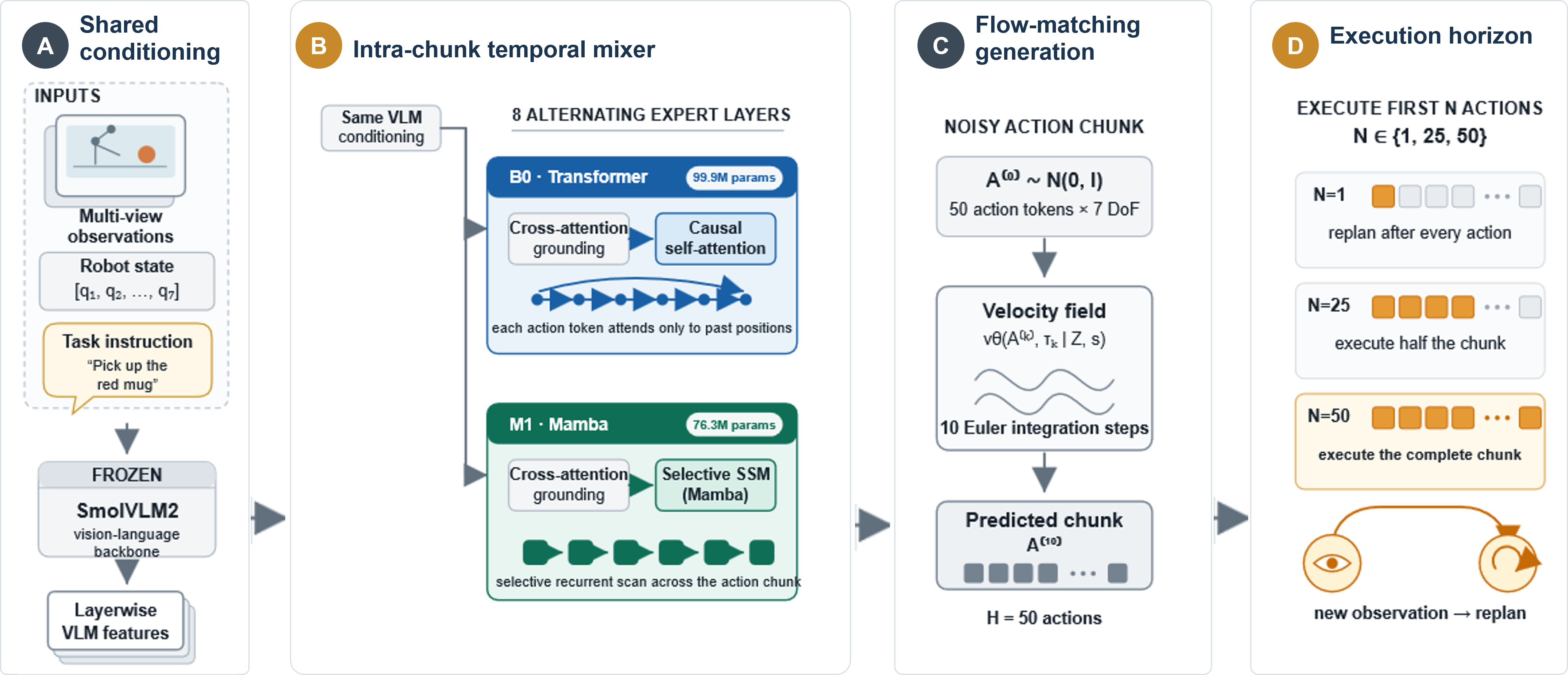}
\caption{Proposed Mamba vs. Transformer Action Expert (AE) pipelines. A) The multi-view RGB observations, the robot states, and the task
instructions are jointly encoded by a SmolVLM2 backbone into layerwise features
that condition the AE. The backbone, its inputs, and the resulting conditioning are identical for both variants. B) The AE alternates
cross-attention grounding layers with intra-chunk temporal layers. The two
variants only differ in the temporal operator: B0 keeps the original Transformer-based causal
self-attention, in which each action token attends to earlier positions of the
chunk, and M1 replaces it with a Mamba-1 selective state-space scan over the
same positions, significantly reducing the number of trainable weights from $99.9$M to $76.3$M. C) Both variants generate a chunk of $H=50$ actions with $D_a=7$ dimensional action vectors using the same conditional flow-matching sampler. 
D) At execution, only the first $N\in\{1,25,50\}$ actions of the chunk are
applied before a new observation is taken and a new chunk is generated.
Since every component other than the Intra-chunk temporal mixers are kept fixed in B), the differences in success rates across $N$ will be attributed to the Mamba vs. Transformer choice.}
\label{fig:overview}
\label{fig:grapg}
\end{figure*}

In recent years, the use of foundation models has gained huge interest throughout robotic perception, planning, and
decision-making, including scene-conditioned reasoning, mission planning, and
high-level behavior generation~\cite{berman2024missiongpt,guo2024vlmauto,zhang2025pla}.
Vision-language-action (VLA) policies extend this line of work to embodied
control by mapping visual observations, language instructions, and robot state
directly to low-level actions~\cite{openvla,pi0,smolvla}. Most recent visuomotor
and VLA policies do not predict a single command per inference. Rather, they emit a
short chunk of continuous actions~\cite{pi0,act,diffusionpolicy}, which
amortizes the cost of one generative forward pass over a horizon of $H$ control steps and
produces locally coherent motions.

During system deployment, the controller executes only the first $N\leq H$ actions of each generated action sequence before requesting a new observation and generating a new sequence prediction. Hence, the choice of $N$ sets a crucial trade-off between VLA accuracy and latency. Using a small $N$ keeps the control policy accurate and close to the current observation but significantly increases the number of VLA inferences, leading to high latency overheads and jeopardizing real-time system deployment~\cite{hadish2024edge}. On the other hand, using a large $N$ reduces the frequency of policy calls but widens the interval during which the robot acts on a stale observation, potentially leading to inaccurate behavior. This is in-line with prior experiments across several VLA backbones which have reported that fewer policy calls are generally accompanied by lower task success rates~\cite{vlacorrector}.

A number of recent approaches have been proposed to address this trade-off at execution time. Among them, Real-Time Chunking~\cite{rtc} improves asynchronous chunk transitions through inference-time inpainting, and VLA-Corrector~\cite{vlacorrector} monitors execution and truncates a chunk once its actions are no longer considered reliable. Crucially, these emerging methods treat the policy as fixed and add \textit{post-hoc} mechanism to alleviate errors in VLA execution. In contrast, this paper proposes a \textit{different and complementary
approach} by investigating how temporal architectural elements \textit{within} the action expert model affect the VLA's success rate when longer action prefixes $N$ are executed.

To investigate this key question, we select the popular SmolVLA architecture~\cite{smolvla} as our reference backbone model in this work and explore how replacing its Transformer-based processing blocks with Mamba-based blocks~\cite{mamba} would improve the system's accuracy-complexity tradeoff. Using SmolVLA as reference backbone is particularly well-suited in the case of this study as it constitutes a state-of-the-art VLA model in terms of achieving high task success rates while featuring lower compute complexity.

In SmolVLA, a pre-trained
vision-language backbone conditions a flow-matching Action Expert (AE) through
layer-wise cross-attention, and the AE alternates cross-attention layers (which ground action features with the multimodal input representation) with causal
self-attention layers (which exchange information among entries within the
noisy action sequence output). This alternation within the AE between cross- and self-attention enables the direct replacement the intra-chunk self-attention temporal mixers, while keeping fixed the remaining layers. Hence, we replace all causal self-attention layers within SmolVLA's AE with Mamba selective state-space model (SSM) blocks~\cite{mamba}. Fig.~\ref{fig:overview} summarizes our proposed pipeline and shed light on the difference between the two model variants considered in this work.

Indeed, Mamba-based SSMs are a well-suited candidate for replacing Transformer-based self-attention layers since Mamba implements input-dependent state-space dynamics through causal
recurrence~\cite{mamba}. An action chunk is a short, ordered control sequence in
which later commands should remain compatible with earlier ones, so a selective
recurrent representation may supply a useful \textit{inductive bias} for intra-chunk
coordination. The open question is whether using Mamba SSMs \textit{improves the accuracy of later actions within the generated chunk} once they are actually executed.

To address this question, the key contributions of this paper are the following:
\begin{enumerate}
    \item We present a novel study on the impact of intra-chunk temporal mixing within flow-matching VLA AEs. We replace SmolVLA's causal
    self-attention layers with Mamba-based SSM mixers while keeping the
    remaining architecture, flow-matching objective, and optimization
    protocol fixed.

    \item We evaluate both our Mamba-based and the Transformer-based reference across three execution horizons using the standard
    LIBERO benchmark suites. 
    
    \item Crucially, we show that our Mamba-based AE significantly improves the VLA's average success rate when using longer action execution horizons, corresponding to feasible real-time system deployments (e.g., an improvement of $+7.8\%$ for $N=50$ and $+3.7\%$ for $N=25$). 

    \item We also demonstrate that our Mamba-based AE reaches a similar success rate of $\sim76\%$ compared to the Transformer-based variant when $N=1$, while reducing the overall VLA model parameter count by $24\%$ due to the compute-efficient nature of Mamba.

    
\end{enumerate}
  
This paper is organized as follows. Section \ref{relatedwork} overviews the related works. Section \ref{sec:method} introduces our proposed methods. Section \ref{sec:setup} details our experimental setup. Section \ref{sec:results} presents our experimental results and discussions. Finally, conclusions are provided in Section \ref{conclusion}.

\section{Related Work} 


\label{relatedwork}
\subsection{Vision-language-action policies and generative AEs}
VLA models adapt pretrained vision-language representations to robot control
using either discrete action tokens or continuous action generators. OpenVLA
treats actions as tokenized outputs of a large vision-language model
\cite{openvla}, whereas $\pi_0$ and SmolVLA use flow-matching AEs to
generate continuous action sequences \cite{pi0, smolvla}. SmolVLA is
particularly suitable for the present study because its expert alternates
cross-attention grounding with causal temporal mixing. This factorization
allows the temporal operator to be modified without changing how visual and
linguistic information conditions the action sequence.

\subsection{Action chunking and the execution horizon}
Action Chunking with Transformers demonstrated that predicting multiple future
actions can improve the temporal coherence of imitation-learning policies
\cite{act}. Diffusion Policy subsequently modeled action chunks as conditional
generative trajectories and introduced receding-horizon execution, in which
only a prefix of each predicted chunk is applied before the plan is
updated from a new observation \cite{diffusionpolicy}. The length of that
prefix is the variable we sweep in this study. Its introduction established
action generation and action execution as related but separable components of a
chunked policy.
 
A more recent line of work targets the cost of the resulting open-loop
interval directly. Real-time chunking addresses asynchronous inference by
constraining newly generated chunks to remain compatible with actions already
committed for execution \cite{rtc}. Concurrent work monitors latent visual
dynamics during execution and truncates the remaining chunk once persistent
deviation is detected, yielding an event-triggered adaptive horizon
\cite{vlacorrector}. Both keep the underlying policy fixed and intervene at
inference time. Our study operates at the opposite end of the same axis: we
hold the execution mechanism fixed and vary the operator that produces the
chunk, which isolates how much of the horizon sensitivity is attributable to
the architecture itself. The two directions compose, since a policy whose
chunks degrade more slowly gives any execution-time monitor a longer usable
interval before it must intervene.

\subsection{State-space models in robot learning}
Mamba introduced selective, input-dependent state-space dynamics as a
competitive sequence-modeling architecture \cite{mamba}. Several robotic
systems have since used Mamba in different parts of the perception and control
pipeline. RoboMamba integrates a Mamba-based multimodal model with a
pose-prediction head for robotic reasoning and manipulation \cite{robomamba}.
MaIL develops a Mamba encoder-decoder architecture for imitation learning,
while Mamba Policy combines SSM and attention components
within a 3D diffusion policy \cite{mail, mambapolicy}. MTIL uses recurrent
state to summarize observation and action history for temporally ambiguous
tasks \cite{mtil}. X-IL studies a broader modular policy design space in which
Mamba and flow-matching components can be interchanged \cite{xil}. FlowRAM uses
Mamba for multimodal fusion within a region-aware flow-matching policy that
predicts manipulation keyframes \cite{flowram}. In contrast, the present work
keeps the VLA backbone and grounding pathway fixed, replaces  the causal
temporal-mixing layers that operate across action positions, and evaluates the
resulting experts as the execution horizon changes. This isolates intra-chunk
temporal modeling from perception, historical context encoding, and
execution-time correction.

\section{Methods}
\label{sec:method}

\subsection{Problem Formulation}

We consider a language-conditioned manipulation policy that maps the current
robot observation and a natural-language instruction to a chunk of future
actions. At environment step $t$, the input observation to the model is
\begin{equation}
    o_t = \left(\mathcal{I}_t,s_t\right),
\end{equation}
where $\mathcal{I}_t=\{I_t^{(v)}\}_{v=1}^{V}$ denotes the available RGB camera
views and $s_t\in\mathbb{R}^{D_s}$ denotes the robot state. When conditioned on an
instruction $\ell$ describing the desired manipulation task, the policy generates an
action chunk
\begin{equation}
    \mathbf{A}_t =
    \left[a_t,a_{t+1},\ldots,a_{t+H-1}\right]
    \in\mathbb{R}^{H\times D_a}.
\end{equation}
We use the SmolVLA action representation with $H=50$ and $D_a=7$ dimensional output action vectors~\cite{smolvla}. Each action
contains a three-dimensional end-effector translation increment, a
three-dimensional rotation increment, and a scalar gripper command.

Our reference baseline model is SmolVLA, which combines a pretrained vision-language backbone with a conditional flow-matching action
expert. The frozen backbone encodes the camera views, robot state, and
instruction into layerwise and multimodal features:
\begin{equation}
    \mathbf{Z}_t
    =
    f_{\phi}(o_t,\ell)
    =
    \left\{\mathbf{Z}_t^{(j)}\right\}_{j=1}^{L_Z},
    \label{eq:backbone}
\end{equation}
where $\phi$ remains fixed during policy training and $L_Z$ denotes the number
of retained backbone layers. These features provide the keys and values used
by the cross-attention layers of the AE within the VLM.

\subsection{Conditional Flow-Matching AE}

Let $(o_t,\ell,\mathbf{A}_t)\sim\mathcal{D}$ denote a training example containing
a demonstrated action chunk, and let
$\boldsymbol{\epsilon}\sim\mathcal{N}(\mathbf{0},\mathbf{I})$. We parameterize
the interpolation from data at $\tau=0$ to noise at $\tau=1$ as:
\begin{equation}
    \mathbf{A}_t^{\tau}
    =
    (1-\tau)\mathbf{A}_t
    +
    \tau\boldsymbol{\epsilon},
    \qquad
    \tau\in[0,1].
    \label{eq:interpolation}
\end{equation}
The endpoints are therefore:
$\mathbf{A}_t^{0}=\mathbf{A}_t$ and
$\mathbf{A}_t^{1}=\boldsymbol{\epsilon}$. The target velocity along this path is
\begin{equation}
    \mathbf{u}_{\tau}
    =
    \frac{\partial\mathbf{A}_t^{\tau}}{\partial\tau}
    =
    \boldsymbol{\epsilon}-\mathbf{A}_t.
\end{equation}
The AE $v_{\theta}$ is trained by minimizing:
\begin{equation}
    \mathcal{L}_{\mathrm{FM}}
    =
    \mathbb{E}_{\substack{
        (o_t,\ell,\mathbf{A}_t)\sim\mathcal{D},\\
        \tau\sim p(\tau),\,
        \boldsymbol{\epsilon}\sim\mathcal{N}(\mathbf{0},\mathbf{I})
    }}
    \left[
        \left\|
        v_{\theta}\!\left(
            \mathbf{A}_t^{\tau},
            \tau,
            \mathbf{Z}_t
        \right)
        -
        \mathbf{u}_{\tau}
        \right\|_2^2
    \right],
    \label{eq:flow_matching}
\end{equation}
where $p(\tau)$ is the original SmolVLA flow-time sampling distribution and is
shared by both the baseline and our Mamba-based variant. At inference, sampling begins from
$\mathbf{A}_t^{1}\sim\mathcal{N}(\mathbf{0},\mathbf{I})$, and the learned
velocity field is integrated from $\tau=1$ to $\tau=0$ using ten Euler steps to
produce the predicted action chunk.

\subsection{Mamba-Based Intra-Chunk Temporal Mixing}

The SmolVLA AE alternates cross-attention grounding layers with
causal self-attention layers. We denote the baseline Transformer expert by
\textbf{B0}. Our variant, \textbf{M1}, retains all eight cross-attention layers
and solely replaces the eight causal self-attention layers with temporal mixers
based on the original Mamba SSM.
(Mamba-1)~\cite{mamba}. Here, M1 denotes our experimental model variant
and should not be confused with the Mamba-1 architecture version. Other than the cross-attention layers, the
vision-language backbone; cross-attention grounding pathway; action and
flow-time embeddings; action-token width; chunk length; and output head remain
unchanged between B0 and M1. 

Let:
\begin{equation}
    \mathbf{X}^{(k)}
    =
    \left[
        \mathbf{x}_1^{(k)},
        \ldots,
        \mathbf{x}_H^{(k)}
    \right]
    \in\mathbb{R}^{H\times d}
\end{equation}
denote the sequence of action-token representations entering temporal layer
$k$, where $d$ is the action-token dimension. The core Mamba-1 selective scan
processes the chunk positions $i=1,\ldots,H$ according to
\begin{align}
    \mathbf{h}_i^{(k)}
    &=
    \overline{\mathbf{A}}_i^{(k)}
    \mathbf{h}_{i-1}^{(k)}
    +
    \overline{\mathbf{B}}_i^{(k)}
    \mathbf{x}_i^{(k)},
    \label{eq:ssm_state}\\
    \mathbf{y}_i^{(k)}
    &=
    \mathbf{C}_i^{(k)}
    \mathbf{h}_i^{(k)}
    +
    \mathbf{D}^{(k)}
    \mathbf{x}_i^{(k)},
    \label{eq:ssm_output}
\end{align}
with $\mathbf{h}_0^{(k)}=\mathbf{0}$. In Mamba-1, the continuous state
transition parameters are learned and shared across sequence positions,
whereas the discretization step $\Delta_i^{(k)}$ and the projections
$\mathbf{B}_i^{(k)}$ and $\mathbf{C}_i^{(k)}$ depend on the current token
representation. Consequently, the discretized transition
$\overline{\mathbf{A}}_i^{(k)}$ depends on
$\Delta_i^{(k)}$, while
$\overline{\mathbf{B}}_i^{(k)}$ depends on both
$\Delta_i^{(k)}$ and $\mathbf{B}_i^{(k)}$. This selectivity allows the mixer
to regulate how information is retained, updated, or suppressed across the
ordered positions of the action chunk.


\subsection{Execution Horizon}

Both experts predict a fixed chunk of $H=50$ actions. At inference, the execution horizon $N$ specifies how many actions are applied before the policy incorporates a new observation and generates another chunk:
\begin{equation}
    \left[a_t,\ldots,a_{t+N-1}\right]
    = \operatorname{First}_{N}(\mathbf{A}_t),
    \qquad N\in\{1,25,50\}.
\end{equation}
At $N=1$, the policy replans after every executed action. At $N=50$, it executes the complete predicted chunk before replanning. The trained weights are held fixed throughout the horizon sweep. Varying $N$ therefore tests how task success changes as increasingly long prefixes of the same type of predicted chunk are used without observation-conditioned correction.

\section{Experimental Setup}
\label{sec:setup}

\subsection{Implementation Details}

Our experimental setup is built using the popular LeRobot framework~\cite{lerobot} in \textit{Python}. The vision-language backbone is
SmolVLM2-500M-Video-Instruct~\cite{smolvlm}, of which we retain the first 16
language-model layers, so $L_Z=16$ in \eqref{eq:backbone}. Input images are
resized to $512\times512$ to match the backbone resolution. The backbone is
initialized from its pretrained weights and frozen throughout learning, such that our training setup updates
the AE alone. The expert stacks eight cross-attention grounding layers
and eight intra-chunk temporal layers and predicts chunks of $H=50$
seven-dimensional delta end-effector actions under the conditional flow-matching
objective in \eqref{eq:flow_matching}. B0 is the unmodified SmolVLA expert, whose
temporal layers are causal self-attention. M1 replaces those eight layers with
Mamba SSM mixers and leaves the grounding layers untouched, significantly reducing the trainable expert from $99.9$M to $76.3$M parameters (a reduction of $23.6\%$). The frozen backbone is identical in both variants and is excluded from
this count.

Both variants are trained for $30{,}000$ updates at a global batch size of 64
using AdamW ($\beta_1=0.9$, $\beta_2=0.95$) with a cosine schedule that warms up
over $1{,}000$ steps and decays the learning rate from $1\times10^{-4}$ to
$2.5\times10^{-6}$. At inference the flow-matching sampler is fixed to ten Euler
steps, and the execution horizon is swept over $N\in\{1,25,50\}$.

\subsection{Benchmark and Protocol}

We evaluate B0 and M1 on the widely-used LIBERO multi-task simulation
benchmark~\cite{libero}, which underpins the evaluation of many recent
VLA policies~\cite{openvla,pi0,smolvla,octo,cotvla}. LIBERO organizes visuomotor
manipulation into four suites, Spatial, Object, Goal, and Long, each containing
ten tasks for a total of 40, chosen to isolate distinct generalization axes:
spatial arrangement, object identity, task goal, and long-horizon composition.
Each task runs in robosuite with a Franka Panda arm~\cite{robosuite}. We follow
the standard LIBERO evaluation protocol. Both variants are trained on the
released demonstration set of $1{,}693$ episodes covering all 40
tasks~\cite{libero,fast}\footnote{LIBERO dataset:
\url{https://huggingface.co/datasets/physical-intelligence/libero}}, and each
task is evaluated over ten rollout trials.

\subsection{Evaluation Metrics}

The primary metric is task success rate. We compute success separately for each
task and multiple model training seeds, then average across the ten tasks within each suite. Reported suite values are means and standard deviations across three seeds. For each
horizon we treat the 40 tasks as paired observations, average each task over the
three seeds, and report the mean per-task difference between M1 and B0, as well as the
$95\%$ confidence interval derived from $10{,}000$ bootstrap resamples of the tasks.

\begin{figure}[t]
\centering
\includegraphics[width=\columnwidth]{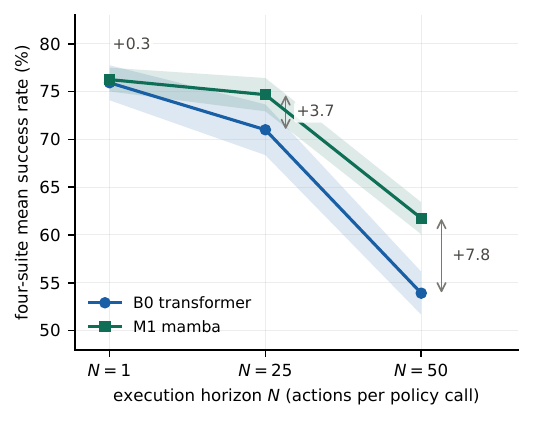}
\caption{Four-suite mean success rate as a function of the execution
horizon $N$, mean across three  seeds with $\pm1$ standard deviation
bands. Annotations give the M1$-$B0 gap at each horizon. }
\label{fig:retention}
\end{figure}

For model $m\in\{\mathrm{B0},\mathrm{M1}\}$, sensitivity to reduced replanning is
summarized by:
\begin{equation}
    \Delta_{\mathrm{drop}}^{m}
    = \operatorname{SR}_{m}(N{=}1)
    - \operatorname{SR}_{m}(N{=}50),
    \label{eq:drop}
\end{equation}
where a smaller value indicates stronger success retention, and the
between-model difference at a given horizon is:
\begin{equation}
    \Delta_{\mathrm{M1-B0}}(N)
    = \operatorname{SR}_{\mathrm{M1}}(N)
    - \operatorname{SR}_{\mathrm{B0}}(N).
    \label{eq:gap}
\end{equation}

\section{Experimental Results}
\label{sec:results}

\subsection{Impact of the Execution Horizon on the Success Rates}

Table~\ref{tab:horizon} and Fig.~\ref{fig:retention} show the four-suite mean
success rate at each horizon. The two experts are separated most clearly under
full-chunk execution. At $N=50$, M1 reaches $61.7\%$ against $53.9\%$ for B0, a significant boost of $7.8$ points over the 40 tasks. For $N=25$, a boost of $3.7$ points is observed, and $0.3$ points at $N=1$. Under per-action replanning (where $N=1$), the two mixers achieves on-par results, and the difference emerges only as the policy is asked to execute longer prefixes without new observations.

Moreover, Moving from $N=1$ to $N=50$, B0 loses $22.0$ points and M1 loses $14.5$, so M1
better retains $7.5$ additional points when the policy must execute a complete predicted
chunk before receiving new visual feedback (Table~\ref{tab:horizon}). M1 obtains this retention with a smaller AE which $76.3$M weights against $99.9$M trainable parameters for B0.

\begin{figure*}[]
\centering
\includegraphics[width=\textwidth]{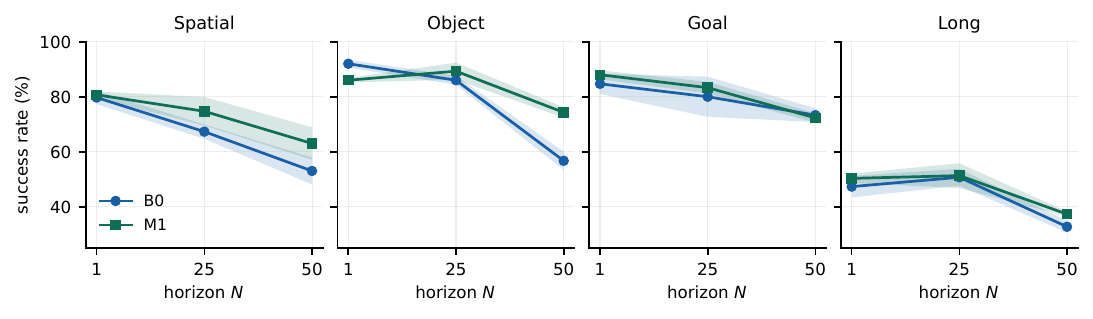}
\caption{LIBERO success rate as a function of execution horizon for B0 and M1,
per suite. Curves show means across three  seeds and shaded bands show
$\pm1$ standard deviation across seeds.}
\label{fig:persuite}
\end{figure*}

\begin{table}[t]
\caption{Four-suite mean success rate (\%) as a function of execution horizon,
mean $\pm$ standard deviation across  seeds. The final column gives the mean per-task difference with a 95\% bootstrap confidence interval over the 40 tasks. }
\label{tab:horizon}
\centering
\begin{tabular}{lccc}
\toprule
Horizon & B0 & M1 & M1$-$B0 [95\% CI] \\
\midrule
$N=1$  & $75.9 \pm 1.8$ & $\mathbf{76.2 \pm 1.2}$ & $+0.3$ $[-3.0, +3.6]$ \\
$N=25$ & $71.0 \pm 2.7$ & $\mathbf{74.7 \pm 1.7}$ & $+3.7$ $[\phantom{-}0.0, +7.3]$ \\
$N=50$ & $53.9 \pm 2.3$ & $\mathbf{61.8 \pm 1.7}$ & $+7.8$ $[+3.0, +12.6]$ \\
\midrule
$N{=}1 \rightarrow 50$ drop & 22.0 & \textbf{14.5} & 7.5 less \\
\bottomrule
\end{tabular}
\end{table}

\begin{table*}[]
\caption{Per-suite success rate (\%) across all three execution horizons,
mean $\pm$ standard deviation over three seeds.}
\label{tab:grid}
\centering
\begin{tabular}{lcccccc c}
\toprule
& \multicolumn{2}{c}{$N=1$} & \multicolumn{2}{c}{$N=25$} & \multicolumn{2}{c}{$N=50$} & gap\\
\cmidrule(lr){2-3}\cmidrule(lr){4-5}\cmidrule(lr){6-7}
Suite & B0 & M1 & B0 & M1 & B0 & M1 & @\,$N{=}50$ \\
\midrule
Spatial & $79.7\pm2.4$ & $80.7\pm1.2$ & $67.3\pm2.6$ & $74.7\pm5.2$ & $53.0\pm4.9$ & $63.0\pm5.9$ & $+10.0$ \\
Object  & $92.0\pm1.4$ & $86.0\pm0.8$ & $86.0\pm1.6$ & $89.3\pm3.1$ & $56.7\pm3.3$ & $74.3\pm1.9$ & $+17.7$ \\
Goal    & $84.7\pm3.7$ & $88.0\pm1.6$ & $80.0\pm7.3$ & $83.3\pm2.1$ & $73.3\pm2.5$ & $72.3\pm1.7$ & $-1.0$ \\
Long    & $47.3\pm3.9$ & $50.3\pm1.7$ & $50.7\pm3.1$ & $51.3\pm4.5$ & $32.7\pm2.1$ & $37.3\pm0.9$ & $+4.7$ \\
\midrule
Average & 75.9 & 76.2 & 71.0 & 74.7 & 53.9 & 61.8 & $+7.8$ \\
\bottomrule
\end{tabular}
\end{table*}

\subsection{Per-Suite Structure}

The aggregate advantage at $N=50$ is not uniform across suites, as shown in Table~\ref{tab:grid} and Fig.~\ref{fig:persuite}. M1 significantly outperforms B0 by $17.7$ points on \texttt{Object} and by $10.0$ points on Spatial, with a more moderate $4.7$-point gain on \texttt{Long}.
\texttt{Goal} is the exception, where B0 leads by a less significant $1.0$ point gap. \texttt{Object} shows the clearest horizon-dependent reversal. B0 leads by $6.0$ points under per-action replanning, $92.0\%$ against $86.0\%$, and trails by $17.7$
points at $N=50$. It is also important to note that \texttt{Long} is the hardest suite for both experts at every horizon, and the lead of M1 over B0 between the two mixers on \texttt{Long} is stable but small relative to the overall average performance gains of M1 over B0.

\subsection{Discussion on the results}
\label{sec:discussion}

This study isolates the effect of the intra-chunk temporal mixer in a flow-matching VLA AE. The Mamba-based variant M1 exhibits stronger success retention at longer open-loop execution horizons, outperforming B0 by an average 3.7 percentage points at $N=25$ and 7.8 points at $N=50$, while the two variants perform comparably at $N=1$ (with M1 further reducing model weight complexity by $24\%$). The longer-horizon settings are practically relevant because executing more actions from each generated chunk reduces the scheduled policy-call frequency, although it also delays the incorporation of new observations.



A plausible interpretation for the clear superiority of our M1 model over B0 is that Mamba's input-dependent causal recurrence provides a useful \textit{inductive bias} for coordinating ordered action positions within the chunk \cite{mamba}. This is in contrast the Transformer-based variant which imposes temporal directionality in the Transformer mixer through an explicit causal attention mask \cite{smolvla}. This architectural distinction may help M1 preserve coordination among later actions when larger sections $N$ of the predicted chunk are executed before replanning.

\section{Conclusion}
\label{conclusion}
This paper has provided a novel investigation of Mamba-based SMM modeling as an alternative to Transformer-based causal self-attention for intra-chunk temporal mixing in flow-matching VLA action experts, and how it affects
task success as implementation-feasible long chunk prefixes are executed. Using a SmolVLA-based experimental setup, it has been demonstrated that the Mamba-based expert retains task success more effectively when using longer open-loop execution horizons, thereby enabling a better accuracy-complexity tradeoff for real-time system deployment. Specifically, under full-chunk execution $N = 50$, our proposed method significantly outperforms the Transformer baseline by $7.8\%$ while featuring a reduction of $\sim 24\%$ in terms of model weight complexity. Our SmolVLA-based experiments demonstrate that success rate at longer execution horizons not only depends on observation staleness, but also on how temporal dependencies within the action chunk are modeled. Future work will examine this behavior across additional VLA architectures, and validate the approach on physical robots.


\end{document}